\documentclass[letterpaper, 10pt, conference]{IEEEtran}
\usepackage[utf8]{inputenc}
\usepackage[T1]{fontenc}
\usepackage{times}
\usepackage{graphicx}
\usepackage{amsmath,amssymb}
\usepackage{booktabs}
\usepackage{hyperref}

\usepackage{cuted}     
\usepackage{capt-of} 

\IEEEoverridecommandlockouts

\title{\LARGE \bf \texttt{roto 2.0}: The Robot Tactile Olympiad}
\author{
    \IEEEauthorblockN{\normalsize
        Elle Miller$^{1}$, Jayaram Reddy$^{2}$, Ayush Deshmukh$^{1}$, Trevor McInroe$^{1}$,
        David Abel$^{1}$, Oisin Mac Aodha$^{1}$, Sethu Vijayakumar$^{1}$
    }
    \thanks{$^{1}$University of Edinburgh, UK. Email: {\tt\small elle.miller@ed.ac.uk}}
    \thanks{$^{2}$National University Singapore, Singapore}
}

\begin{document}

\maketitle 

\begin{strip}
    \centering
    \vspace{-2cm}
    \includegraphics[width=\textwidth]{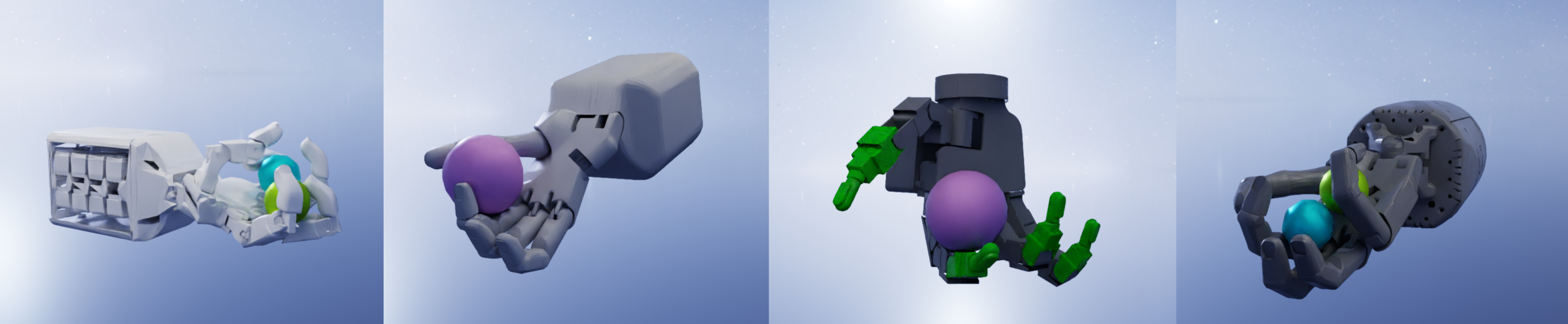}
    \captionof{figure}{\textbf{The \texttt{roto 2.0} Benchmark Suite:} A standardised RL framework across four distinct dexterous morphologies (from L to R): ORCA Hand, Shadow Lite, Allegro Hand, and the Shadow Dexterous Hand. The suite facilitates ``blind" tactile manipulation tasks, such as Baoding ball rotation and ball bouncing.}
    \label{fig:banner}
\end{strip}

\begin{abstract}
Tactile-based reinforcement learning (RL) is currently hindered by fragmented research and a focus on over-saturated orientation tasks. We introduce v2 of the Robot Tactile Olympiad (\texttt{roto 2.0}), a GPU-parallelised benchmark designed to standardise tactile-based RL across four distinct robotic morphologies (16-DOF to 24-DOF). Unlike prior benchmarks, \textbf{\texttt{roto}} focuses on end-to-end ``blind'' manipulation, utilising only proprioception and tactile sensing without state information or distillation.
We demonstrate a significant performance leap, with our blind agents achieving 13 Baoding ball rotations in 10 seconds, an order of magnitude faster than current state-of-the-art speeds. By open-sourcing our environments and robustly tuned baselines, we reduce the barrier to entry and enable researchers to prioritise fundamental algorithmic challenges over tedious RL tuning. Website: \href{https://elle-miller.github.io/roto/}{https://elle-miller.github.io/roto/}
\end{abstract}

\section{Introduction}

Real-world robotic manipulation requires the ability to interact robustly in unstructured environments where visual line-of-sight is frequently obstructed. To achieve this, robots must learn to ``feel.'' While Reinforcement Learning (RL) has revolutionised locomotion across complex terrains \cite{rudin2025parkourwildlearninggeneral}, tactile-based manipulation lags significantly behind. Progress is currently hindered by a fragmented landscape: most labs work in isolation using unique combinations of sensors and robots, making cross-validation difficult. Furthermore, the community has largely over-saturated the task of in-hand orientation, e.g. 
\cite{andrychowicz_learning_2020, sievers_learning_2022, chen_system_nodate, qi_-hand_2022, qi_general_2023, rostel_estimator-coupled_2023, yang2024anyrotategravityinvariantinhandobject, suresh_neural_2023}. While impressive, a single task fails to capture the broader spectrum of challenges that tactile observations pose, leaving the true utility of tactile feedback an open debate.
The difficulty of tactile-based RL stems from a trifecta of complexity: manipulation is inherently hard \cite{mason_toward_2018}, on-policy RL is notoriously difficult to tune, and thus effectively combining the two with sparse and discontinuous tactile observations is a significant undertaking. This difficulty is exacerbated by a lack of standardised tactile-based RL benchmark environments across complex tasks and morphologies. \textit{Tactile-Gym 2.0} \cite{tactilegym} is limited to a 4-DOF arm and single-point contact tasks, while  \textit{VTDexManip} \cite{liu2025vtdexmanip} focuses on pre-training policies with pre-collected datasets.
To fill this gap, we introduce v2 of the Robot Tactile Olympiad (\textbf{\texttt{roto 2.0}}), an RL benchmark built upon GPU-parallelised Isaac Lab \cite{isaaclab}. Originally introduced in \cite{miller2025tactilerl} for the Shadow Hand, we expand the suite here to include four distinct dexterous morphologies: the anthropomorphic Shadow Dexterous Hand (24-DOF), Shadow Dexterous Hand Lite (16-DOF), Allegro Hand (16-DOF) and ORCA Hand (17-DOF).
Crucially, we demonstrate that with our RL training pipeline, ``blind'' policies (utilising only proprioception and tactile data) can master sophisticated manipulation without the need for the teacher-student distillation or explicit pose estimators common in prior work \cite{sievers_learning_2022, yin_rotating_2023, yang_tacgnn_2023}. While the current state-of-the-art for Baoding ball rotation achieves a maximum of 3 rotations in 10 seconds \cite{yuan_robot_2023}, our blind agents exploit the raw potential of tactile-proprioceptive loops to achieve significantly higher throughput of up to 15 rotations. As demonstrated in \cite{miller2025tactilerl}, integrating self-supervised forward dynamics to aid representation learning can further push this boundary to 25 rotations per 10 seconds, nearly closing the gap between blind policies and state-based agents.
From these results, we argue that establishing robust haptic foundations is a prerequisite to integrating visual modalities, rather than a secondary addition to vision-centric systems. By open-sourcing \textbf{\texttt{roto}}, we aim to reduce the barrier-to-entry for researchers interested in tactile-based RL and help focus community efforts on high-impact research directions instead of RL tuning. 
Contributions:
\begin{itemize}
   \item \textbf{The \texttt{roto} benchmark:} A set of tactile-based environments with integrated hyperparameter optimisation and robustly tuned baselines.
\item \textbf{Morphological diversity:} A cross-platform evaluation of four robot hands to study how hardware complexity influences tactile policy convergence.
\item \textbf{Performance breakthroughs:} We provide ``blind'' policies that achieve state-of-the-art speeds in complex manipulation, providing a new ceiling for tactile intelligence.
\end{itemize}

\section{Methodology}

\textbf{RL.} We use a customised implementation
of Proximal Policy Optimisation (PPO) \cite{schulman_proximal_2017} from SKRL~\cite{serrano2023skrl} to incorporate observation stacking, self-supervision, separated environments for continuous evaluation, and various training tricks \cite{miller2026theartofrobotrl}. We use $8,092$ parallelised environments for training and $100$ for evaluation. For each combination of robot ($n=3$), task ($n=2$), and observation setting (blind vs. state-based), we perform a hyperparameter sweep across seven PPO parameters using $40$ trials ($8$ warm-up runs) to ensure robust baselines.

\begin{itemize}
    \item \textit{Bounce:} The agent must bounce a ball as many times as possible in 10 seconds (600 timesteps). A bounce is defined as a contact event after a period of at least 5 timesteps ($\sim 83$ms) without contact. 
    \item \textit{Baoding:} The agent must rotate two balls (55g) around each other in-hand as many times as possible within 10 seconds (600 timesteps). We use 1.5 inch diameter for Shadow Hand and ORCA, 2 inches for Allegro and 1.2 inches for Shadow Lite.
\end{itemize}
\textbf{MDP.} The blind agents receive a history length $k=4$ of proprioceptive and binary tactile observations, and are joint-position controlled, see Table~\ref{tab:obs} for details. We define the each task-relevant robot link as a tactile sensor. The state-based agents additionally receive the object position(s) and linear velocity.
For \emph{Bounce}, the agent is rewarded with  $r_{bounce}=10$ for every successful bounce. For \emph{Baoding}, we specify two static target positions and define the reward as $r_{{dist}_1} + r_{{dist}_2} + r_{rotation}$. The dense distance rewards $r_{{dist}_1}, r_{{dist}_2}$ encourage the balls to the targets. When the centers of both balls are within 1.0 cm of the targets, the targets switch and the agent receives a bonus reward $r_{rotation}=10$. The episode terminates if any object is out of reach or the maximum episode length $T=600$ is reached. The physics simulation runs at 240 Hz, the control policy at 60 Hz.

\begin{table}[h!]
    \centering
    \caption{Observation and action spaces}
    \resizebox{\columnwidth}{!}{
    \begin{tabular}{llccccc}
    \hline
          Type  & Description & \emph{Shadow} & \emph{Shadow Lite} & \emph{Allegro} & \emph{ORCA}  \\
         \hline
           Tactile obs. & binary contacts  & 17 & 14 & 20 & 17 \\
           \hline
        
         Proprio. obs. & joint positions & 20 & 16 & 16 & 17 \\
          & joint velocities & 20 & 16 & 16 & 17\\
         & joint command error & 20 & 13 & 16 & 17\\
        & last action & 20 & 13 & 16 & 17\\
         \hline
         Total obs. & single timestep & 97 & 72 & 84 & 85 \\
         & $k=4$ timesteps & 388  & 288 & 336 & 340 \\
         \hline
         Actions & joint positions & 20 & 13 & 10 & 17 \\
         \hline
    \end{tabular}}
    \label{tab:obs}
\end{table}

\section{Experimental results}

\begin{figure}[t!]
    \centering
    \includegraphics[width=0.75\linewidth]{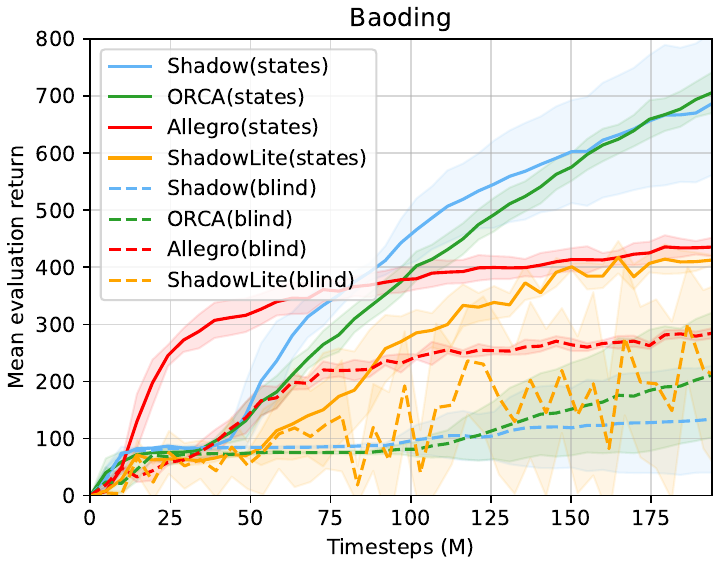}
    \includegraphics[width=0.75\linewidth]{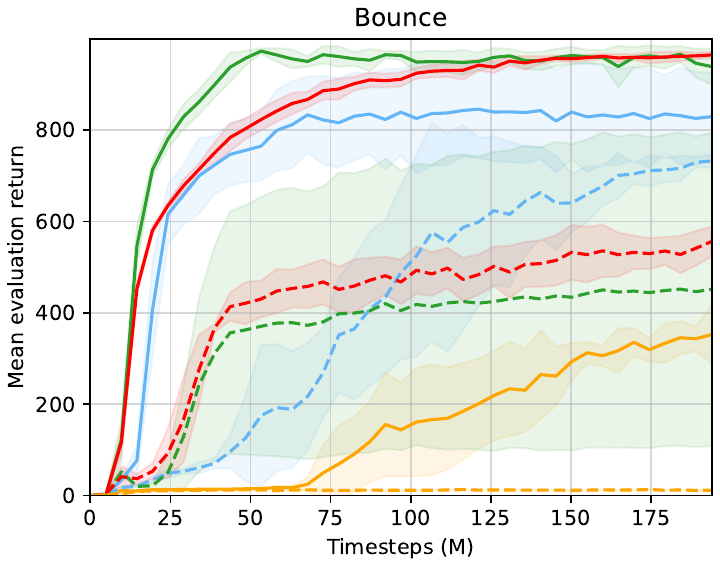}
    \caption{Mean evaluation returns across 5 seeds for state-based and blind agents in the Baoding and Bounce tasks.}
    \label{fig:results}
    \vspace{-0.5cm}
\end{figure}

We evaluate the learning efficiency and asymptotic performance of four distinct morphologies for state-based and blind agents. The mean evaluation returns are summarised in Figure~\ref{fig:results}; we refer the reader to the project page for the policy videos. In the Bounce environment, state-based agents approach the theoretical maximum return ($1,000$ reward, corresponding to $100$ successful bounces). While the hands converge to similar success rates, they exhibit hardware-specific strategies, e.g. the ORCA hand adopts an ``outstretched'' starfish-like pose. Our blind agents demonstrate high sample efficiency, approaching $80$ bounces by $200$M steps. Despite the vast differences in hardware, we find that performance trends remain consistent across the full-hand morphologies, excepting the Shadow Lite.
The Baoding task reveals a more significant performance gap between state-based and blind agents. State-based agents achieve a throughput of up to $35$ rotations in $10$ seconds. Blind agents exhibit much lower performance with high variance; while a top-performing Shadow Hand seed achieved $13$ rotations, others failed to converge. This stochasticity highlights a core challenge in tactile-based RL: efficient feature extraction \cite{miller2025tactilerl}.

\section{Discussion \& Conclusion}

We introduce \textbf{\texttt{roto 2.0}}, a multi-morphology benchmark for blind dexterous manipulation. While our high-speed simulated policies represent a ``performance ceiling'' that exceeds current real-world hardware limits, they provide a benchmark for developing the next generation of robust RL pipelines.
Our results indicate that while blind policies can approach privileged performance in simple tasks like bouncing, complex or multi-object manipulation (Baoding) remains an open challenge. We identify high-priority research directions for the community: beyond sparse binary contacts to richer forms of tactile information, ML methodologies with inductive biases for tactile data, and expanding tasks beyond hands e.g. whole-body humanoid manipulation \cite{sferrazza2024humanoidbench}. 
We are actively investigating the sim-to-real transferability of these policies and welcome community contributions to the \textbf{\texttt{roto}} suite to accelerate the arrival of the ``locomotion moment" for robotic touch.

\bibliographystyle{IEEEtran}
\bibliography{references}

\end{document}